\documentclass[runningheads]{llncs}
\usepackage[T1]{fontenc}
\usepackage{cite}
\usepackage{amsmath,amssymb,amsfonts}
\usepackage{algorithmic}
\usepackage{textcomp}
\usepackage{booktabs}
\usepackage{arydshln}
\usepackage{subfig}
\usepackage{siunitx}
\usepackage{caption}
\usepackage{multirow}
 \usepackage{xspace}
 \usepackage[utf8]{inputenc}
\usepackage[table,svgnames]{xcolor} 

\usepackage{graphicx}
\begin{document}
\title{Seeing More with Less:\\
Video Capsule Endoscopy with Multi-Task Learning}
%

%\titlerunning{Advancing Video Capsule Endoscopy: A Multi-Task Learning Approach.}
\author{Julia Werner\inst{1}\orcidID{0009-0006-0279-1776} \and
Oliver Bause\inst{1}\orcidID{0009-0003-5388-2959} \and
Julius Oexle\inst{1} \and
Maxime Le Floch\inst{2,3} \and Franz Brinkmann\inst{2,3}\orcidID{0000-0002-3474-3115} \and Jochen Hampe\inst{2,3}\orcidID{0000-0002-2421-6127} \and Oliver Bringmann\inst{1}\orcidID{0000-0002-1615-507X}}
\authorrunning{J. Werner et al.}
\institute{Department of Computer Science, University of Tübingen, Tübingen,
Germany \and
Else Kröner Fresenius Center for Digital Health, TU Dresden, Dresden, Germany \and
 Department of Medicine I, University Hospital Dresden, TU Dresden, Dresden, Germany }

%\email{Placeholder}\\ \and
%Placeholder\\}
%
\maketitle              % typeset the header of the contribution
\begin{abstract}
Video capsule endoscopy has become increasingly important for investigating the small intestine within the gastrointestinal tract. However, a persistent challenge remains the short battery lifetime of such compact sensor edge devices. 
Integrating artificial intelligence can help overcome this limitation by enabling intelligent real-time decision-making, thereby reducing the energy consumption and prolonging the battery life.
However, this remains challenging due to data sparsity and the limited resources of the device restricting the overall model size. 
In this work, we introduce a multi-task neural network that combines the functionalities of precise self-localization within the gastrointestinal tract with the ability to detect anomalies in the small intestine within a single model. Throughout the development process, we consistently restricted the total number of parameters to ensure the feasibility to deploy such model in a small capsule. 
We report the first multi-task results using the recently published Galar dataset, integrating established multi-task methods and Viterbi decoding for subsequent time-series analysis. This outperforms current single-task models and represents a significant advance in AI-based approaches in this field.
Our model achieves an accuracy of 93.63\% on the localization task and an accuracy of 87.48\% on the anomaly detection task. 
The approach requires only 1 million parameters while surpassing the current baselines.

\keywords{Video Capsule Endoscopy  \and Multi-Task Learning \and Viterbi decoding.}
\end{abstract}

\section{Introduction}
The Video Capsule Endoscopy (VCE) aims to detect pathological tissue within the gastrointestinal (GI) tract, more specifically the small intestine, by employing a small pill-size capsule, which is equipped among others with LEDs and a camera~\cite{iddan2000wireless,thomson2001small}. 
By peristalsis, the capsule traverses through the GI tract and captures low-resolution images throughout its journey, which are then sent out to an external device for final evaluation by medical doctors. This procedure is essential as it transmits images of the small intestine, which is largely not accessible by standard techniques, such as the gastroscopy or colonoscopy~\cite{costamagna2002prospective,smedsrud2021kvasir}.
Thus, the main objective of the VCE is to cover the small intestine, while the esophagus, stomach and colon are of less interest.
%The state-of-the-art devices only capture images, but detailed evaluation is performed by medical doctors afterwards.

Current devices, such as the PillCam SB3~\cite{pillcam}, simply transmit all captured images to an external device without prior classification.
However, this transmission is very costly and only those frames originating from the small intestine are actually relevant in the VCE. 
On the other hand, simply storing all images on the device locally until the end of the procedure requires collection of the capsule afterwards~\cite{zwinger2019capsocam}, does not improve the battery lifetime and prohibits real-time assessment as well as decision-making.
Importantly, as the available energy of such small sensor edge device is very limited, e.g. the PillCam has a battery lifetime of 8-12h~\cite{pillcam,monteiro2016pillcam}, it is essential to minimize unnecessary energy usage.
Since the traversal time of the capsule varies between patients, for some patients, the small intestine is not covered by such capsule, before the battery is depleted.
Theoretically, images that are not of interest can be discarded immediately, saving the energy that would be otherwise be used for their transmission.
To achieve this, the organ, which the capsule currently traverses, needs to be classified and transmission of images only started after exiting the stomach.\\
\hspace*{1.8em}Furthermore, jointly addressing organ detection with anomaly detection while continuously limiting the overall model size, can improve the entire procedure by ensuring comprehensive coverage of the GI tract and enabling preliminary assessments of potential anomalies prior to final evaluation by physicians. 
On-site anomaly detection allows real-time decisions to be made, such as raising the frame rate or resolution, which might improve the visualization of important regions.
However, AI-based anomaly detection still remains a major challenge in this field due to data sparsity and the need of high medical competence to label such data. One possible technique to tackle this is multi-task learning (MTL)~\cite{caruana1997multitask} which leverages the observation that, in certain cases, simultaneous learning of multiple tasks can produce superior results~\cite{caruana1997multitask,zhang2021survey}. This approach relies on the assumption that the tasks being targeted are related. When this condition is met, MTL has the potential to surpass the performance of single-task learning (STL).\\
\hspace*{1.8em}\textbf{Our Contribution: } By applying MTL to VCE, our main objective is to develop a model capable of concurrently classifying the organ section and assessing whether an anomaly is present in the captured image. 
Precisely determining the capsule's entry into the small intestine enables suppression of image transmission outside the target region and thus, reduces energy consumption.
This helps to ensure full small intestine coverage across all patients without exhausting the battery. 
Thus, we report the first multi-task results targeting this localization problem along with anomaly detection and aim to surpass the performance of current state-of-the-art single-task models for both objectives. 
Since for real-world applications, this model must be deployed on a very small device, its hardware implementation feasibility is a key consideration. As an initial step, we focus on limiting the overall model size and the total number of multiply-accumulate (MAC) operations compared to the current baselines.
\section{Related Work}

\paragraph{Multi-Task Studies Targeting the Gastrointestinal Tract:}
Recently, several MTL approaches have been developed to capture diseases in the GI tract.
For example, \cite{kong2021multi} targeted the detection of Crohn's disease, by using MTL with a private dataset including standard endoscopy images from 15 patients, however, using very large networks, such as a DenseNet121 and a ResNet50, for this task.
More research has been conducted to detect esophageal lesions~\cite{yu2021multi} and polyps in the colon \cite{tang2023transformer} with MTL.
Notably, both studies did not target the small intestine. 
Most recently, \cite{xu2024multi} performed magnetically controlled capsule endoscopy for the classification of gastric anatomical sites as well as lesions on a public, self-built dataset.
In this work, following the constraints of VCE applications, only datasets containing low-resolution images from VCE studies with a specific focus on targeting the small intestine are applicable.

\paragraph{Localization and Anomaly Detection with VCE Images:}
The reported results are predominantly limited to single-task models, which can either identify the anatomical regions or classify lesions in the digestive tract. This is inherently given by the fact that the previously published VCE datasets contain labels only for a single task. For example, the Kvasir-Capsule dataset~\cite{smedsrud2021kvasir} consists only of images labeled for anomalies but not the organ sections while the Rhode-Island Gastroenterology dataset~\cite{charoen2022rhode} comprises of images labeled for the different organ sections but not for any anomalies. 
The newly available large Galar dataset~\cite{le2025galar} incorporates images labeled not only for different GI sections but also anomalies. 
Besides \cite{le2025galar}, \cite{werner2025enhanced} were the first to perform anomaly detection on the Galar dataset with an ensemble model, however; without targeting the localization task. 
We intend to perform MTL on VCE images drawn from this dataset and outperform current single-task baselines by combining the localization and the anomaly detection task while restricting the overall model size. Since the small intestine is the only part of the GI tract largely inaccessible by standard endoscopes, it is the primary focus of this procedure.

\section{Experiments and Methodology}

\begin{figure}[htbp]
		\centering\includegraphics[width=0.98\linewidth]{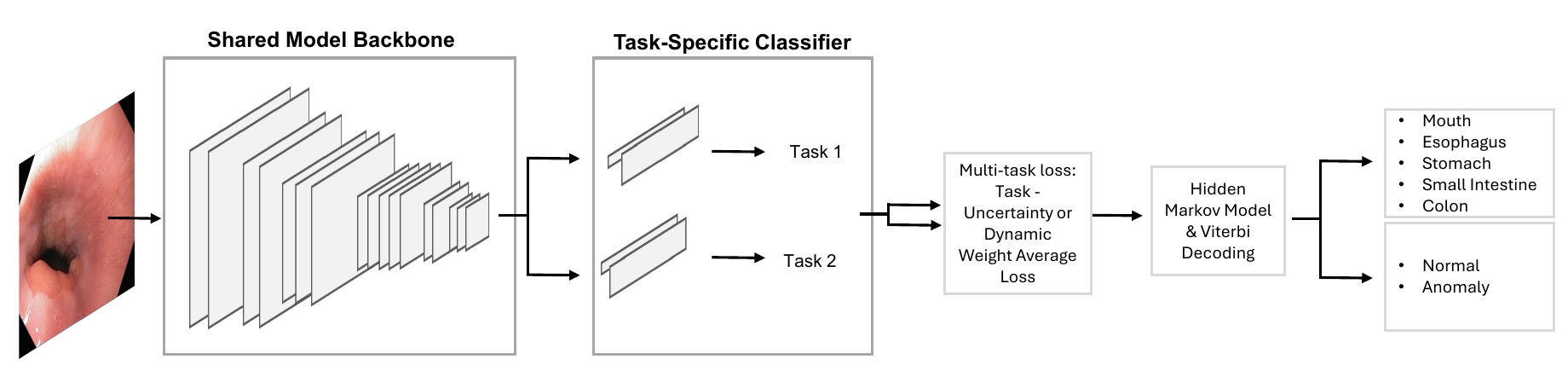}
			\caption{The neural network is trained on the Galar dataset with VCE images and hard parameter sharing with two separate classifier heads, dynamic weight average or task uncertainty weighting of losses and post-processing with a HMM and Viterbi decoding.
				\label{fig:mtl}}
\end{figure}

The primary goal of our method is to prepare a single, resource-constrained, light-weight neural network, that is capable of performing both, localization and anomaly detection tasks, on VCE images as presented in Figure~\ref{fig:mtl} to ensure coverage of the small intestine before the capsule's battery is depleted and provide basic anomaly detection functionalities. 
We start by preparing the dataset to capture class imbalances, continue with the neural network training including hard-parameter sharing, and finally perform post-processing for the localization task involving a HMM and Viterbi decoding.

\subsection{Pre-Processing - Galar Dataset}

To the best of our knowledge, the Galar dataset~\cite{le2025galar} is the only publicly available VCE dataset, which consists of images labeled for the different GI organs as well as anomalies. Thus, this dataset is well-suited for simulating the VCE environment and evaluating multi-task learning approaches for this application. 
The authors of the Galar dataset selected patients with the ID $61-80$ for the test set, representing completely untouched VCE studies from a clinical setting. For better comparability, we evaluate our method on the same test set. The remaining $60$ patient studies (ID $1-60$) were used as a development set for training the neural network.

As multiple other medical datasets, the Galar dataset is characterized by a large class imbalance. From all organs available in this dataset, the smallest class (mouth) represents only $\SI{0.05}{\percent}$ of all images. 
Moreover, pathological cases represent the minority within the classes, and in addition, there are significant differences in class frequencies among the pathologies themselves. While the smallest anomaly class (erythema) comprises only $\SI{0.12}{\percent}$ of the data, the largest class (blood) makes up $\SI{11.9}{\percent}$ of all frames.
Additionally, if the capsule traveled slowly or resides at one location for some time, many similar images are often captured, which can disrupt the training process. 
To counteract this class imbalance, the majority classes were downsampled as described in the following.

%Since there are many frames that depict multiple pathologies at once, the larger classes still account for a larger fraction but are in total drastically reduced. For example, the downsampling leads to a reduction of $242,000$ to $22,000$ blood samples. 
The main objective remains to theoretically capture all images of interest and transmit them to an on-body device for final evaluation by physicians.
We do not aim to provide an extensive evaluation of anomalies on-site, but to capture optimally all relevant images, which are finally evaluated by medical doctors after transmission.
Additionally, basic on-site anomaly detection allows important features to be added, such as increased resolution or frame rate upon outlier detection improving the entire procedure.
Thus, detailed analysis of different anomaly types does not need to be conducted on the capsule directly.
Instead, using a binary distinction eases the computational workload of the model for this application and also enables size restrictions to be applied.
Therefore, all pathologies were combined into a single anomaly class in order to perform binary classification. For each pathology, a random selection of frames was included. %- up to five times the number in the smallest class - was included. 
Next, all samples from the mouth and esophagus were collected and the remaining section classes randomly downsampled to have a final ratio of $1:1$ (normal : anomaly). 
The data was randomly shuffled and split into a training and validation set with a defined ratio of $70:30$.
The final class distribution of the different organ sections and the anomaly and normal samples is depicted in Table~\ref{tab:class_distribution}.

\begin{table}[htbp]
\caption{Class distribution of the used organ sections and anomalies (positive samples) after downsampling to address the occurrence of many related frames.}
\begin{center}
\resizebox{\textwidth}{!}{%
\begin{tabular}{l@{\hskip 0.2cm} | c@{\hskip 0.2cm} c@{\hskip 0.2cm} c@{\hskip 0.2cm} c@{\hskip 0.2cm} c@{\hskip 0.2cm} c@{\hskip 0.2cm} | c@{\hskip 0.2cm} c@{\hskip 0.2cm} c@{\hskip 0.2cm}}
\toprule
Subset & Mouth & Esophagus & Stomach & Small Intestine & Colon & Total & Negative & Positive & Total \\\midrule\midrule
Train           & 1,476                   & 1,718                    & 11,962         & 32,274  & 26,059    & 73,489 & 39,779           & 40,104    & 79,883    \\ 
Val         & 252                    & 132                           & 2,598      & 9,478 & 7,633   & 20,093 &  10,748                    &      10,757  &  21,505 \\ 
Test     &    265    &    397    &   16,510 & 175,716 & 41,667 & 234,555 & 218,665                   & 16,499  & 235,164  \\\bottomrule 
\end{tabular}%
}
\end{center}
\label{tab:class_distribution}
\end{table}

\subsection{Neural Network Training with Hard-Parameter Sharing}
In the context of capsule endoscopy, multiple authors~\cite{xu2024multi,werner2023precise} aimed to restrict the total number of parameters to address the constraints given by tiny edge devices equipped during such procedure by using light-weight networks from the MobileNet family. These networks are not only beneficial in image classification tasks but were additionally designed for the efficient deployment on embedded devices~\cite{sandler2018mobilenetv2,howard2019searching}. 
Hence, in this work, we adopt the MobileNetV3-Small~\cite{howard2019searching} architecture and modify the classification head. Instead of a single classification head, we designed and generated separate classification heads tailored to each specific task. Offering the benefit of minimizing parameters and straightforward deployment while allowing the classification of multiple tasks, we implement hard parameter sharing~\cite{caruana1993multitask}. This MTL technique is used to share the hidden layers between all tasks and further exploit task-specific classifier heads. 

The pretrained model was fine-tuned for additional $10$ epochs and the learning rate and weight decay were each defined based on a hyperparameter search with Hydra~\cite{Yadan2019Hydra} and the well-established hyperparameter optimization framework Optuna~\cite{akiba2019optuna} with $30$ trials (for the localization, the anomaly detection and the multi-task runs each). We aimed to employ well-established MTL losses for comparison and to investigate how this influences the final results.
Thus, the homoscedastic task uncertainty based on~\cite{kendall2018multi} was implemented. The MTL loss $\mathcal{L}$ is assembled for the tasks $i \in \{1,2\}$, with 
\begin{align}
    \mathcal{L} = \sum_{i=1}^n\exp{(-\log\sigma_i)} \mathcal{L}_i +\log\sigma_i.
\end{align} 

%Abschnitt Julius

Furthermore, the Dynamic Weight Average Loss (DWA) \cite{liu2019end} was implemented to improve the balancing of the loss values for both tasks.
Instead of weighting the tasks according to their homoscedastic uncertainty, this loss adjusts its weights based on loss changes between the epochs. This helps to prevent premature neglect of any task and supports the learning of more challenging tasks over extended training periods. The temperature factor $T$ also allows the strength of the weighting to be set manually. The DWA weighting $\lambda_i$ is assembled for the tasks $i \in \{1,2\}$, with 
\begin{align}
    \lambda_i(t):=\frac{I \exp{(w_i(t-1)/T)}}{\sum_k \exp{(w_k(t-1)/T)}},
    w_i(t-1)=\frac{\mathcal{L}_i(t-1)}{\mathcal{L}_i(t-2)}.
\end{align}

For the anomaly detection task, the cross-entropy loss was replaced with a focal loss \cite{lin2017focal} to address the pronounced class imbalance between normal and abnormal tissue samples. This allows the model to focus more on the underrepresented class without requiring oversampling of abnormal cases, thereby preserving a realistic class distribution during training.
The final evaluations were then passed to a HMM to perform post-processing.
%hier noch Formel für Focal Loss?

%Abschnitt Julius

\subsection{Post-Processing - Hidden Markov Model and Viterbi Decoding}

The HMM~\cite{rabiner1989tutorial} has previously been applied to the problem of localization in video capsule endoscopy, using the only available dataset at the time with labels for 4 organ sections: the Rhode Island VCE dataset~\cite{charoen2022rhode}~\cite{werner2023precise}. 
It has been shown, that specifically for this setting involving time-series data, post-processing the CNN output with a HMM and Viterbi decoding leads to a significant improvement in performance allowing a more precise determination of when the small intestine is entered compared to solely using a CNN.
This method is particularly well-suited to the problem at hand as it incorporates prior knowledge of the sequential order in which the organs are traversed and their respective lengths.

\begin{figure}[htpb]
\centering
    \includegraphics[scale=0.45]{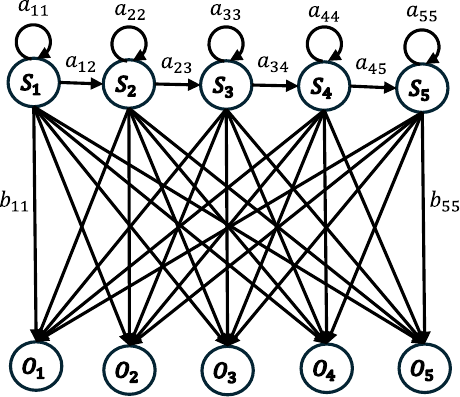}
    \caption{Hidden Markov Model for the described GI setting.}
    \label{fig:hmm}
\end{figure}

In this work, we aim to transfer this technique to the newly available Galar dataset that includes class labels for the organs across the GI tract and apply it as a post-processing to the MTL approach. Compared to~\cite{werner2023precise}, the number of hidden states is extended to $5$, such that we have 
$\{S_1=\mathrm{Mouth}, S_2=\mathrm{Esophagus}, S_3=\mathrm{Stomach}, S_4=\mathrm{Small \ intestine}, S_5=\mathrm{Colon}\}$,
 with an initial distribution $\pi_i=P(X_1=S_i)$ for $i\le 5$.The HMM for this GI setting is visualized in Figure~\ref{fig:hmm}.
 
Each moment $t$, at which the capsule captured an image, an observation $X_t \in {K_1, \dots, K_m}$ is obtained, representing a location classified by the neural network. The transition probabilities $a_{ij}=P(X_{t+1} =S_j | X_t = S_i)$ of the Markov chain $S_t$ are encoded such that for $S_{t+1}$ only the current organ or the succeeding organ can be the next state, in accordance with the natural anatomy of the GI tract. The emission probabilities $b_{j}(k) = P(O_t = K_k | X_t = S_j )$ for an observation $O_t \in \{O_1=\mathrm{Mouth}, O_2=\mathrm{Esophagus}, O_3=\mathrm{Stomach}, O_4=\mathrm{Small \ intestine}, O_5=\mathrm{Colon}\}$ at time $t$ are inferred based on the confusion matrix from the neural network on the training set as this directly encodes the emission probabilities. 
Finally, the Viterbi algorithm~\cite{forney1973viterbi} calculates the most probable sequence of gastroenterological states $X=(X_1,\dots, X_t)$, based on the evaluations of the neural network $(O_1,\dots,O_t)$, with $\underset{X_1,\dots,X_{t}}{\arg\max} \: P(X_1,\dots, X_{t}, O_1, ... O_{t})$ as described in~\cite{werner2023precise}.% with a sliding window size of $300$.

\section{Results}
Our main objective is to provide a very light-weight multi-task model, that enhances the current baselines of single-task models, while reducing the overall complexity and provide useful functionalities for VCE devices, predominantly by ensuring the coverage of the whole small intestine by precise organ detection/localization of the capsule.
Typically, deep neural networks are too large to be deployed on tiny edge devices, such as the video capsule for endoscopies. To address this, we conducted all experiments using the light-weight MobileNet~\cite{howard2019searching}.  
First, in Table~\ref{tab:results_mt_localization}, the results for the organ detection are presented.

% 63,614,128 st loc macs 
% 63,611,056 st anomaly macs
% 63,616,176 mt focal dwa and dwa und uw
\begin{table}[htbp]\centering\tiny
\caption{Results of the multi-task (MT) runs for the \textbf{localization} task in [$\si{\percent}$] with the \textbf{MobileNetV3} compared to single-task (ST) baseline results with the best results in bold and the second best results underlined.}
\resizebox{\textwidth}{!}{%
\begin{tabular}{lcccccc}\\\toprule
  & Accuracy & F1 & Precision & Recall &  Params & MACs \\\midrule\midrule
Baseline~\cite{le2025galar} & 81 & 71 & -- & -- & 25 M & 4,087 B\\
Our ST Localization & 85.05 & 60.55 & 54.42 & 86.44 & 1 M & 63,614 M\\\hline\hline 
Our MT Results \\ \hline\hline
MT UW  w/o HMM & 80.96 & 49.30 & 48.50 & 69.58 & 1 M & 63,616 M\\
MT DWA w/o HMM & 86.22 & 61.53 & 54.48 & 87.08 & 1 M & 63,616 M\\
MT DWA Focal w/o HMM & 87.77 & 65.12 & 58.61 & 87.5 & 1 M & 63,616 M\\
MT UW \& HMM & 90.48 & 66.22 & 71.57 & 79.14 & 1 M & 63,616 M\\
MT DWA \& HMM & \underline{93.48} & \underline{91.02} & \underline{88.56} & \underline{94.34} & 1 M & 63,616 M \\
MT DWA Focal \& HMM & \textbf{93.63} & \textbf{92.41} & \textbf{90.40} & \textbf{94.94} & 1 M & 63,616 M\\
\bottomrule 
\end{tabular}\label{tab:results_mt_localization}%
}
\end{table}

The ST model shows inferior results compared to the baseline with a F1-score of only $60.55\%$ possibly due to a less complex model, however, this is notably improved by the MT approach.
The proposed MT approach, if combined with Viterbi decoding, shows a strong performance across all tested loss functions.
The best results were achieved with the DWA and focal loss, with an accuracy of $93.63\%$ and a F1-score of $92.41$ compared to an accuracy of $81\%$ and F1-score of $71\%$ of the baseline~\cite{le2025galar}.
We further observe, that the uncertainty weighted loss shows inferior results compared to the DWA or DWA with focal loss.

Furthermore, in Table~\ref{tab:results_mt_anomaly}, the MT results for the anomaly detection task are shown.
While all three loss options lead to proficient results, the DWA loss produces the best results compared to the baseline and the ST model (F1-score: 54.38\% vs. 37.01\%).
Only the recall is inferior compared to the baseline, with only 54.69\% vs. 60.88\%. 
However, this work provides a VCE targeted approach to perform the important tasks of localization and anomaly detection within one single model.
Notably, both tasks can be conducted with the same model of only 1 M parameters, which is only $25\%$ of the anomaly detection baseline model and merely $4\%$ of the localization baseline model.
In addition, the number of MAC units is reduced from 4 B and 189 M to only 64 M with this approach, lowering the computational workload.

\begin{table}[htbp]\centering\tiny
\caption{Results of the multi-task (MT) runs for the \textbf{anomaly detection} task in [$\si{\percent}$] with the \textbf{MobileNetV3} compared to single-task (ST) baseline results.}
\resizebox{\textwidth}{!}{%
\begin{tabular}{lcccccc}\\\toprule
Classification  & Accuracy & F1 & Precision & Recall &  Params & MACs \\\midrule\midrule
Baseline~\cite{werner2025enhanced} & 87.28 & 37.01 & 26.59 &\textbf{60.88} & 4 M & 189 M\\
Our ST Anomaly Detection & 84.91 & \underline{53.34} & 52.99 & 54.63 & 1 M & 63,611 M\\\hline\hline
Our MT Results \\ \hline\hline
MT UW  & 83.69 & 52.94 & 52.71 & 54.66 & 1 M & 63,616 M\\
MT DWA & \textbf{87.7} & \textbf{54.38} & \textbf{54.14} & \underline{54.69} & 1 M & 63,616 M\\
MT DWA Focal & \underline{87.48} & 53.24 & \underline{53.08} & 53.45 & 1 M & 63,616 M \\
\bottomrule 
\end{tabular}\label{tab:results_mt_anomaly}%
}
\end{table}

%Additionally, for the first patient from the test set (ID=61) the classification differences between the MT CNN alone and the MT CNN combined with the HMM are depicted in Figure~\ref{fig:hmm_eval}. It can be seen that the Viterbi Decoding detects many misclassifications and outperforms the CNN if used in combination with the MT model.

%\begin{figure}[hptb]
 %   \centering
  %  \subfloat[placeholder]{\includegraphics[width=0.5\textwidth]{figures/output_all.pdf}\label{fig:avg_energy}}
   % \subfloat[Classification performance of the \textbf{MT CNN $+$ HMM} (last row). First detection of the small intestine marked in red.]{\includegraphics[width=0.5\textwidth]{figures/output_all.pdf}\label{fig:hmm_eval}}
    %\caption{}
    %\label{fig:grid_search}
%\end{figure}

% first multi-task model for these tasks in capsule endoscopy
% first with this dataset
% restriction of model size -consideration of mac units less than 4M as in related work
% first combined with hmm multtitask in this setting
% better localization accuracy than galar

\section{Conclusion}
This work introduces a light-weight neural network with a total of just 1 M parameters, designed to combine precise self-localization and anomaly detection capabilities through a multi-task learning approach for VCE. We report the first multi-task results using the recently published Galar dataset and integrate various established multi-task methods and Viterbi decoding. With an accuracy of $\SI{93.63}{\percent}$ and an F1-score of $\SI{92.41}{\percent}$, our approach shows superior performance for the localization task compared to the current baseline, while also enabling fundamental anomaly detection functionality.
Thus, the presented work improves upon current AI models for VCE, by providing precise localization within the GI tract and basic anomaly detection functionalities.

% uw inferior, if combined with hmm, strong performance, low cost on hardware, pplicable to other low power applicatios involving time-series data. Outperforming the baseline with a smaller model size 
\textbf{Prospect of application:} Once the exit of the stomach has been determined by the multi-task model, additional features such as adapting frame rates (lowered before the small intestine and increased upon anomaly detection) or varying resolutions may be added. The approach can prolong battery life and serve as a starting point for deploying multi-task models on VCE devices.

%Use maximum 60 words to describe the prospect and envisioned contexts, scenarios, or circumstances on the potential application/deployment of your work. 

%\begin{credits}
%\subsubsection{\ackname} This study was partially funded
%by X (grant number Y).
%\end{credits}

\subsubsection{Acknowledgments} This work has been partly funded by the German Federal Ministry of Research, Technology and Space (BMFTR) in the project MEDGE (16ME0530).
\subsubsection{Disclosure of Interests.}
%It is now necessary to declare any competing interests or to specifically
%state that the authors have no competing interests. Please place the
%statement with a bold run-in heading in small font size beneath the
%(optional) acknowledgments\footnote{If EquinOCS, our proceedings submission
%system, is used, then the disclaimer can be provided directly in the system.},
%for example: 
The authors have no competing interests to declare that are
relevant to the content of this article. 

\bibliographystyle{splncs04}
\bibliography{mybibliography}

\end{document}